\documentclass[a4paper]{article}

\usepackage{INTERSPEECH_v2}
\usepackage{array}
\usepackage{subfig}

\title{Residual LSTM: Design of a Deep Recurrent Architecture for Distant Speech Recognition}

\name{Jaeyoung~Kim$^1$, Mostafa~El-Khamy$^1$, Jungwon~Lee$^1$}
\address{
  $^1$Samsung Semiconductor, Inc. \\
  4921 Directors Place, San Diego, CA, USA}
\email{jaey1.kim@samsung.com, mostafa.e@samsung.com, jungwon2.lee@samsung.com}

\begin{document}

\maketitle
\begin{abstract}
In this paper, a novel architecture for a deep recurrent neural network,
residual LSTM is introduced. A plain LSTM has an internal memory cell that can
learn long term dependencies of sequential data. It also
provides a temporal shortcut path to avoid vanishing or exploding gradients in the temporal domain.
The residual LSTM provides an additional spatial shortcut
path from lower layers for efficient training of deep networks with multiple LSTM
layers. Compared with the previous work, highway LSTM, residual LSTM separates
a spatial shortcut path with temporal one by using output layers, which
can help to avoid a conflict between spatial and temporal-domain gradient flows.
Furthermore, residual LSTM reuses the output
projection matrix and the output gate of LSTM to control the spatial information
flow instead of additional gate networks, which effectively reduces more than 10\%
of network parameters.
An experiment for distant speech recognition on the AMI SDM corpus shows
that 10-layer plain and highway LSTM networks presented 13.7\%
and 6.2\% increase in WER over 3-layer baselines, respectively.
On the contrary, 10-layer residual LSTM networks provided the lowest WER 41.0\%, which
corresponds to 3.3\% and 2.8\% WER reduction over plain
and highway LSTM networks, respectively.

\end{abstract}

\section{Introduction}
Over the past years, the emergence of deep neural networks has fundamentally
changed the design of automatic speech recognition (ASR). Neural network-based
acoustic models presented significant performance improvement over the prior
state-of-the-art Gaussian mixture model (GMM)~\cite{dahl2012contextTT,
hinton2012deep, lecun1995convolutional, lecun1998gradient, dahl2011large}. Advanced neural
network-based architectures further improved ASR performance. For example,
convolutional neural networks (CNN) which has been huge success in image
classification and detection were effective to reduce environmental and
speaker variability in acoustic features~\cite{sainath2015convolutional,
swietojanski2014convolutional, abdel2012applying, sainath2013deep, abdel2014convolutional}.
Recurrent neural networks
(RNN) were successfully applied to learn long term dependencies of sequential data
~\cite{graves2013hybrid, sak2014long, sak2014long2, schuster1997bidirectional}.

The recent success of a neural network based architecture mainly comes from its
deep architecture~\cite{morgan2012deep,srivastava2015training}.
However, training a deep neural network is
a difficult problem due to vanishing or exploding gradients. Furthermore,
increasing depth in recurrent architectures such as gated recurrent unit (GRU)
and long short-term memory (LSTM) is significantly more
difficult because they already have a deep architecture in the temporal domain.

There have been two successful architectures for a deep feed-forward neural
network: residual network and highway network.
Residual
network~\cite{he2015deep} was successfully applied to train more than 100 convolutional
layers for image classification and detection.
The key insight in the residual network is to provide a shortcut path between layers that
can be used for an additional gradient path.
Highway network~\cite{srivastava2015highway} is
an another way of implementing a shortcut path in a feed-forward neural
network. \cite{srivastava2015highway} presented successful
MNIST training results with 100 layers.

Shortcut paths have also been investigated for RNN and LSTM networks. The maximum entropy
RNN (ME-RNN) model~\cite{mikolov2011strategies} has direct connections between the input and
output layers of an RNN layer. Although limited to RNN networks with a single hidden layer,
the perplexity improved by training the direct connections as part of the whole network.
Highway LSTM~\cite{zhang2016highway,yao2015depth} presented a multi-layer
extension of an advanced RNN architecture, LSTM~\cite{hochreiter1997long}.
LSTM has internal memory cells that provide shortcut gradient paths in the temporal direction.
Highway LSTM reused them for a highway shortcut in the spatial domain.
It also introduced new gate networks to control highway paths from the prior layer
memory cells. \cite{zhang2016highway} presented highway LSTM for far-field speech recognition and
showed improvement over plain LSTM. However, \cite{zhang2016highway} also showed
that highway LSTM degraded with increasing depth.

In this paper, a novel highway architecture, residual LSTM is introduced. The key
insights of residual LSTM are summarized as below.
\begin{itemize}
\item Highway connection between output layers instead of internal memory cells: LSTM internal
memory cells are used to deal with gradient issues in the temporal
domain. Reusing it again for the spatial domain could make it more difficult to train a
network in both temporal and spatial domains. The proposed residual LSTM network uses an
output layer
for the spatial shortcut connection instead of an internal memory cell, which can
less interfere with a temporal gradient flow.
\item Each output layer at the residual LSTM network learns residual mapping not
learnable from highway
path. Therefore, each new layer does not need to waste time or resource to generate similar
outputs from prior layers.
\item Residual LSTM reuses an LSTM projection matrix as a gate network. For an usual LSTM
network size, more than 10\% learnable parameters can be saved from residual
LSTM over highway LSTM.
\end{itemize}
The experimental result on the AMI SDM corpus~\cite{carletta2005ami} showed
10-layer plain and highway LSTMs had severe degradation from increased depth:
13.7\% and 6.2\% increase in WER over 3-layer baselines, respectively. 
On the contrary, a 10-layer residual LSTM presented
the lowest WER 41.0\%, which outperformed the best models of plain and highway LSTMs.



\section{Revisiting Highway Networks}
\label{sec:revisit}

In this section, we give a brief review of LSTM and three existing highway architectures.
\subsection{Residual Network}
A residual network~\cite{he2015deep} provides an identity mapping by shortcut paths.
Since the identity mapping is always on, function output only needs to
learn residual mapping. Formulation of this relation can be expressed as:
\begin{equation}
\label{eq1_resnet}
y=F(x;W)+x
\end{equation}
$y$ is an output layer, $x$ is an input layer and $F(x;W)$ is a function with
an internal parameter $W$. Without a shortcut path, $F(x;W)$ should represent $y$ from
input $x$, but with an identity mapping $x$, $F(x;W)$ only needs to learn residual mapping, $y-x$.
As layers are stacked up, if no new residual mapping is needed, a network can
bypass identity mappings without training, which could greatly simplify
training of a deep network.

\subsection{Highway Network}

A highway network~\cite{srivastava2015highway} provides another way of implementing
a shortcut path for a deep neural-network. Layer
output $H(x;W_h)$ is multiplied by a transform gate $T(x;W_T)$ and before going into the
next layer, a highway path $x\cdot(1-T(x;W_T ))$ is added. Formulation of a highway network
can be summarized as:
\begin{equation}
\label{eq1_highwaynetwork}
        y=H(x;W_h)\cdot  T(x;W_T) +x \cdot (1-T(x;W_T)) 
\end{equation}
A transform gate is defined as:
\begin{equation}
\label{eq2_highwaynetwork}
T(x;W_T) =\sigma(W_T x+b_T)
\end{equation}
Unlike a residual network, a highway path of a highway network is not always turned on.
For example, a highway network can ignore a highway path if $T(x;W_T )=1$ , or
bypass a output layer when $T(x;W_T )=0$.

\subsection{Long Short-Term Memory (LSTM)}

Long short-term memory (LSTM)~\cite{hochreiter1997long} was proposed to resolve
vanishing or exploding gradients for a recurrent neural network. LSTM
has an internal memory cell that is controlled by forget and input gate networks.
A forget gate in an LSTM layer determines how much of prior memory value should be passed
into the next time step. Similarly, an input gate scales new input to memory
cells. Depending on the states of both gates, LSTM can represent long-term or short-term
dependency of sequential data. The LSTM formulation is as follows:
\begin{eqnarray}
\label{eq1_lstm}
        i_t^{l} & = &\sigma(W_{xi}^{l} x_t^{l}+W_{hi}^{l} h_{t-1}^{l}
        +w_{ci}^{l} c_{t-1}^{l}+b_i^{l})  \\
\label{eq2_lstm}
        f_t^{l} & = &\sigma(W_{xf}^{l} x_t^{l}+W_{hf}^{l} h_{t-1}^{l}
        +w_{cf}^{l} c_{t-1}^{l}+b_f^{l})  \\
\label{eq3_lstm}
        c_t^{l} & = &f_t^{l} \cdot c_{t-1}^{l}+
    i_t^{l} \cdot \tanh(W_{xc}^{l} x_t^{l}+W_{hc}^{l} h_{t-1}^{l}+b_c^{l}) \\
\label{eq4_lstm}
        o_t^{l} & =& \sigma(W_{xo}^{l} x_t^{l}+W_{ho}^{l} h_{t-1}^{l}
        +W_{co}^{l} c_t^{l}+b_o^{l}) \\
\label{eq5_lstm}
    r_t^l & =& o_t^l \cdot \tanh(c_t^l )\\
\label{eq6_lstm}
    h_t^l & =& W_p^l \cdot r_t^l
\end{eqnarray}
$l$ represents layer index and
$i_t^l$, $f_t^l$ and $o_t^l$ are input, forget and output gates respectively. They
are component-wise multiplied by input, memory cell and hidden output to
gradually open or close their connections. $x_t^l$ is an input from ${(l-1)}^{th}$ layer
(or an input to a network when $l$ is 1), $h_{t-1}^l$ is a $l^{th}$ output layer at time $t-1$ and
$c_{t-1}^{l}$ is an internal cell state at $t-1$.
$W_p^l$ is a projection matrix to reduce dimension of $ r_t^l$.

\subsection{Highway LSTM}
Highway LSTM~\cite{zhang2016highway, hochreiter1997long}
reused LSTM internal memory cells for
spatial domain highway connections between stacked LSTM layers.
Equations (\ref{eq1_lstm}), (\ref{eq2_lstm}), (\ref{eq4_lstm}), (\ref{eq5_lstm}), and
(\ref{eq6_lstm}) do not change for highway LSTM.
Equation (\ref{eq3_lstm}) is updated to add a highway connection:

\begin{eqnarray}
        c_t^l&=& d_t^l \cdot c_t^{l-1}+f_t^l \cdot c_{t-1}^l+\nonumber\\
        &&i_t^l\cdot\tanh(W_{xc}^l x_t^l+W_{hc}^l h_{t-1}^l+b_c^l) \\
       d_t^l&=&\sigma(W_{xd}^l x_t^l+W_{cd}^l c_{t-1}^l+w_{cd}^l c_t^{l-1}+b_d^l)
\end{eqnarray}
Where $d_t^l$ is a depth gate that connects $c_t^{l-1}$ in the $(l-1)^{th}$
layer to $c_t^l$ in the $l^{th}$ layer. \cite{zhang2016highway} showed that
an acoustic model based on the highway LSTM network improved far-field speech recognition
compared with a plain LSTM network. However, ~\cite{zhang2016highway} also showed that word error rate (WER)
degraded when the number of layers in the highway LSTM network increases from 3 to 8.

\section{Residual LSTM}
\label{sec:resLSTM}
In this section, a novel architecture for a deep recurrent neural network,
residual LSTM is introduced.
Residual LSTM starts with an intuition that the separation of a spatial-domain
shortcut path with a temporal-domain cell update may give better flexibility to
deal with vanishing or exploding gradients.
Unlike highway LSTM, residual LSTM does not accumulate a
highway path on an internal memory cell $c_t^l$. Instead, a shortcut path is added to an
LSTM output layer $h_t^l$. For example, $c_t^l$ can keep a temporal gradient
flow without attenuation by maintaining forget gate $f_t^l$ to be close to one.
However, this gradient flow
can directly leak into the next layer $c_t^{l+1}$ for highway LSTM in spite
of their irrelevance. On the contrary, residual LSTM has less impact from $c_t^l$ update due to separation of gradient paths.

\begin{figure}[!t]
\centering
\includegraphics[width=3.5in]{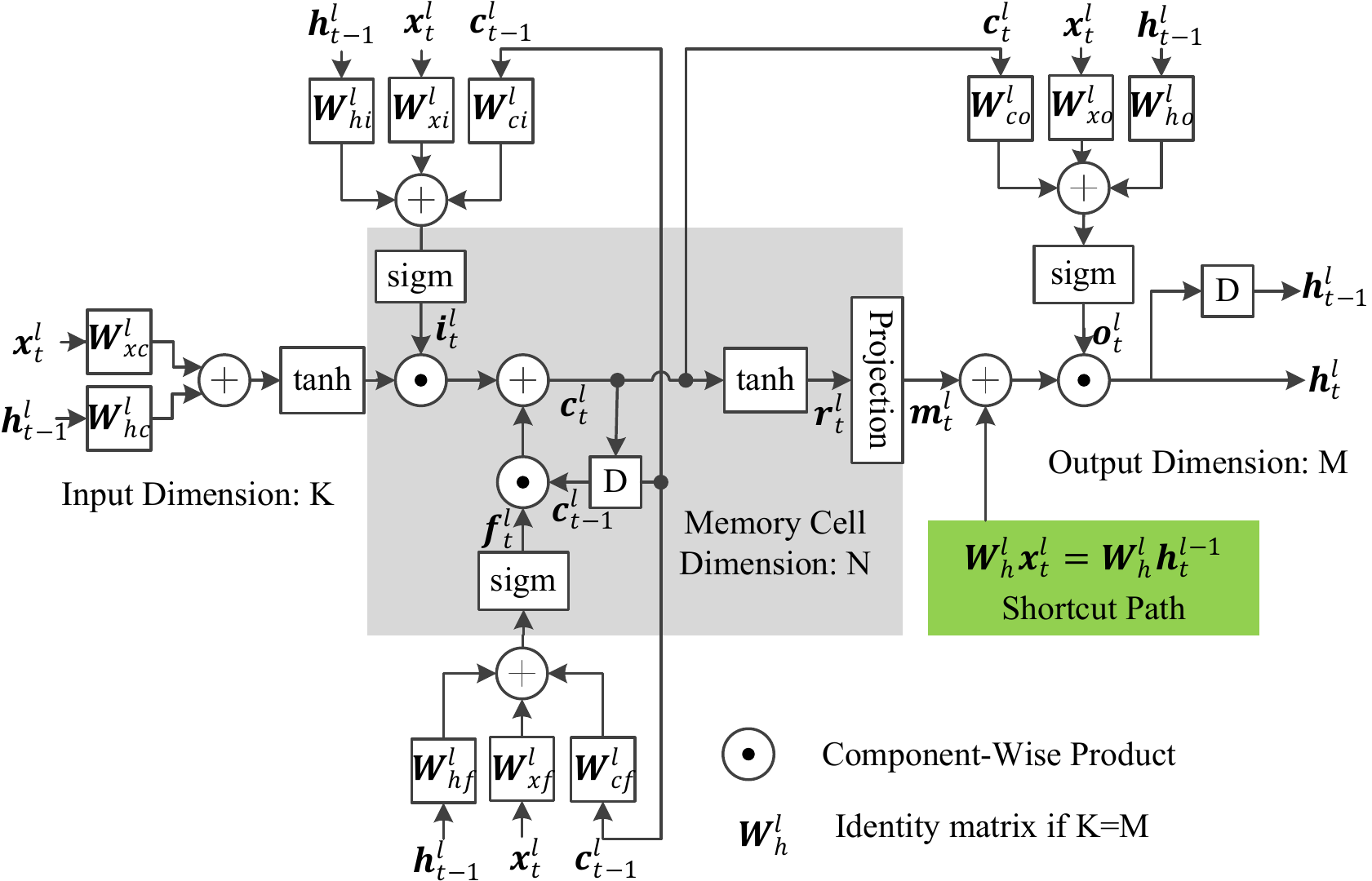}
\caption{Residual LSTM: A shortcut from a prior output layer $h_t^{l-1}$
is added to a projection output $m_t^{l}$. $W_h^{l}$ is a dimension matching matrix
between input and output. If $K$ is equal to $M$, it is replaced with an identity matrix.}
\label{fig:resLSTM_cell}
\end{figure}

Figure~\ref{fig:resLSTM_cell}
describes a cell diagram of a residual
LSTM layer. $h_t^{l-1}$ is a shortcut path from $(l-1)^{th}$ output layer that is added to
a projection output $m_t^l$. Although a shortcut path can be any lower output layer, in this
paper, we used a previous output layer.
Equations (\ref{eq1_lstm}), (\ref{eq2_lstm}), (\ref{eq3_lstm}) and (\ref{eq4_lstm}) do not
change for residual LSTM. The updated equations are as follows:

\begin{eqnarray}
\label{eq5_residuallstm}
        r_t^{l} &= &\tanh(c_t^{l} ) \\
\label{eq6_residuallstm}
        m_t^{l} &= & W_p^{l} \cdot r_t^{l} \\
\label{eq7_residuallstm}
        h_t^{l} & =& o_t^{l} \cdot (m_t^{l}+W_h^{l} x_t^{l})%
\end{eqnarray}

Where $W_h^{l}$ can be replaced by an identity
matrix if the dimension of $x_t^{l}$ matches
that of $h_t^{l}$.
For a matched dimension, Equation~(\ref{eq7_residuallstm}) can be changed into:
\begin{equation}
h_t^{l}  = o_t^{l} \cdot (m_t^{l}+ x_t^{l})%
\end{equation}
Since a highway path is always turned on for residual LSTM, there should be a
scaling parameter on the main path output. For example, linear filters
in the last CNN layer of a residual network are reused to
scale the main path output. For residual LSTM, a projection matrix $W_p^{l}$ is reused in
order to scale the LSTM output.
Consequently, the number of parameters for residual LSTM does not increase compared
with plain LSTM.
Simple complexity comparison between residual LSTM and highway LSTM is as follows.
If the size of the internal memory cells is $N$ and the output
layer dimension after projection is $N/2$, the total number of reduced parameters for a residual
LSTM network
becomes $N^2/2+4N$. For example, if $N$ is 1024 and the number of layers is more than 5,
the residual LSTM network has approximately 10\% less network parameters compared with the highway LSTM network with same $N$ and a projection matrix.

One thing to note is that
a highway path should be scaled by an output gate as in Equation~(\ref{eq7_residuallstm}).
The initial design of residual LSTM was to simply add an input path to an LSTM output without
scaling, which
is similar to a ResLSTM block in~\cite{zhang2016very}.
However, it showed 
significant performance loss
because highway paths keep accumulated as the number of layers increase.
For example,
the first layer output without scaling would be $o_{t}^{1}\cdot m_{t}^{1}+x_{t}^{1}$,
which consists of
two components. For the second layer output, however, the number of components increases as
three: $o_{t}^{2}\cdot m_{t}^{2} + o_{t}^{1}\cdot m_{t}^{1}+x_{t}^{1}$.
Without
proper scaling, the variance of an residual LSTM output will keep increasing.

The convolutional LSTM network proposed in~\cite{zhang2016very} added batch normalization layers, which
can normalize increased output variance from a highway path.
For residual LSTM, output gate is re-used to act similarly
without any additional layer or parameter.
Output gate is a trainable network which can learn a proper range of an LSTM output. For example,
if an output gate is set as $\frac{1}{\sqrt{2}}$, an $l^{th}$ output layer becomes
\begin{equation}
h_t^l=\sum_{k=1}^l (\frac{1}{\sqrt{2}})^{(l-k+1)} m_t^k + (\frac{1}{\sqrt{2}})^{l} x_t
\end{equation}
Where, $x_t$ is an input to LSTM at time $t$. If $m_t^l$ and $x_t$ are independent
each other for all $l$ and have fixed variance of 1, regardless of layer index $l$,
the variance of layer $l^{th}$ output becomes $1$. Since variance of a output layer is variable
in the real scenario, a trainable output gate will better deal with exploding variance
than a fixed scaling factor.

\section{Experiments}
\label{sec:experiment}
\subsection{Experimental Setup}
AMI meeting corpus~\cite{carletta2005ami} is used to train and evaluate residual LSTMs.
AMI corpus consists of 100 hours of meeting recordings. For each
meeting, three to four people have free conversation in English.
Frequently, overlapped speaking from multiple speakers happens and for that
case, the training transcript always follows a main speaker.
Multiple microphones are used to synchronously record conversations in different environments.
Individual headset microphone (IHM) recorded clean close-talking conversation and single
distant microphone (SDM) recorded far-field noisy conversation.
In this paper, SDM is used to train residual LSTMs at
Section~\ref{sec:exp_train} and~\ref{sec:exp_fixlayer}
and combined SDM and IHM corpora are used at Section~\ref{sec:exp_increasedC}.

Kaldi~\cite{Povey_ASRU2011} is a toolkit for speech recognition that is used to
train a context-dependent LDA-MLLT-GMM-HMM system. The trained GMM-HMM generates forced
aligned labels which are later used to train a neural network-based acoustic model.
Three neural network-based acoustic models are trained: plain LSTM network without any
shortcut path, highway LSTM network and residual LSTM network.
All three LSTM networks have 1024 memory cells and 512 output nodes
for experiments at Section~\ref{sec:exp_train}, ~\ref{sec:exp_fixlayer}
and~\ref{sec:exp_increasedC}.

The computational network toolkit (CNTK)~\cite{yu2014introduction} is used to train and decode
three acoustic models.
Truncated back-propagation through time (BPTT) is used to train LSTM
networks with 20 frames for each truncation.
Cross-entropy loss function is used with L2 regularization.

For decoding, reduced 50k-word Fisher dictionary is used for lexicon
and based on this lexicon, tri-gram language model is
interpolated from AMI training transcript. As a decoding option, word
error rate (WER) can be calculated based on non-overlapped speaking or
overlapped speaking.
Recognizing overlapped speaking is to decode up to 4 concurrent speeches.
Decoding overlapped speaking is a big challenge
considering a network is trained to only recognize a main speaker. Following
sections will provide WERs for both options.

\subsection{Training Performance with increasing Depth}
\label{sec:exp_train}

\begin{figure}[!t]
\centering
\subfloat[]
{
\includegraphics[width=2.5in]{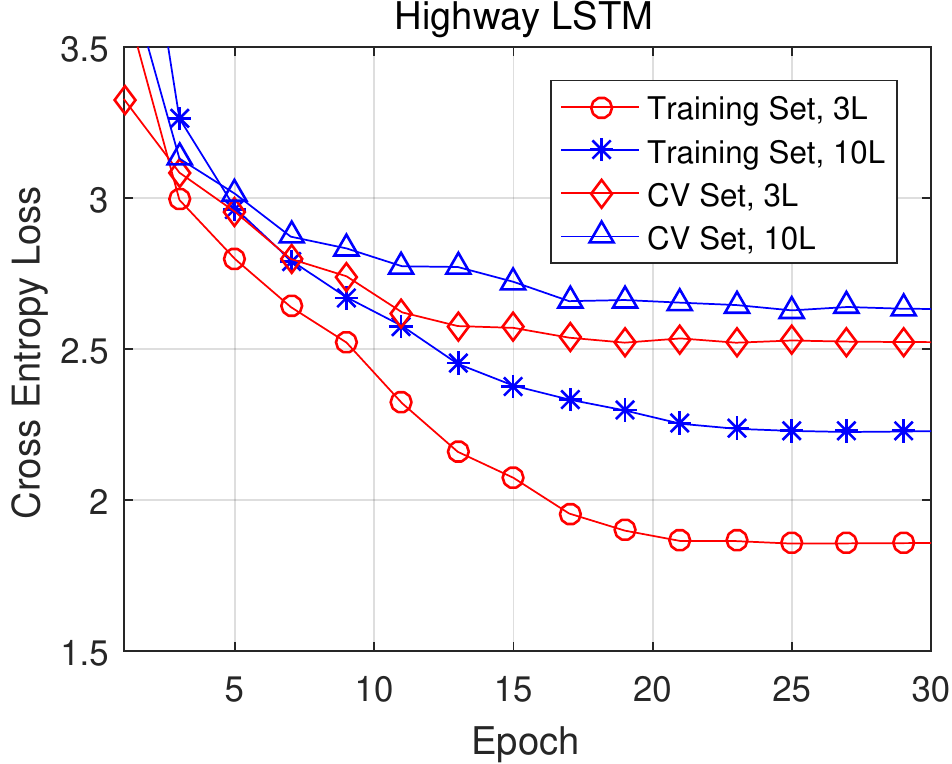}
\label{fig:highway_train}
}
\hfill
\subfloat[]
{
\includegraphics[width=2.5in]{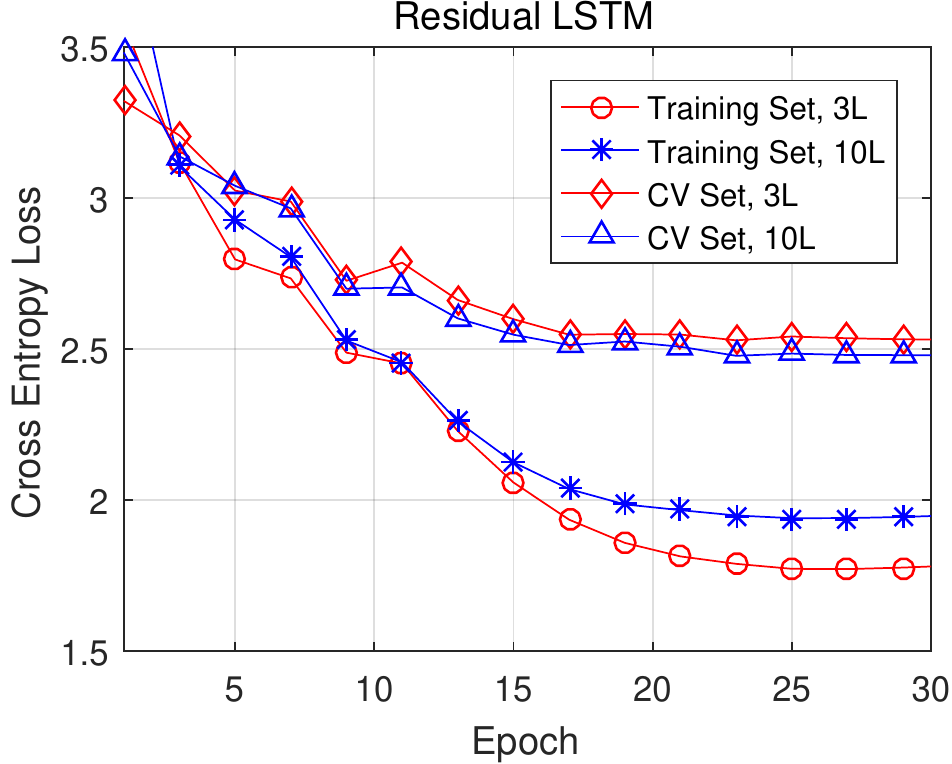}
\label{fig:res_train}
}

\caption{Training and CV PERs
on AMI SDM corpus. (a) shows training and cross-validation (CV) cross-entropies
for 3 and 10-layer highway LSTMs.
(b) shows training and cross-validation (CV) cross-entropies for 3 and 10-layer 
residual LSTMs.
}
\label{fig:train}
\end{figure}

Figure~\ref{fig:train} compares training and cross-validation (CV) cross-entropies
for highway and residual LSTMs. The cross-validation set is only used to evaluate
cross-entropies of trained networks.

In Figure~\ref{fig:highway_train}, training and CV cross-entropies for a 10-layer
highway LSTM increased 15\% and 3.6\% over 3-layer one, respectively. 
3.6\% CV loss for a 10-layer highway LSTM does not come from overfitting because the training
cross-entropy was increased as well. The training loss from increased network depth was observed
in many cases such as Figure 1 of~\cite{he2015deep}. A 10-layer highway LSTM revealed
the similar training loss pattern, which implies highway LSTM
does not completely resolve this issue.

In Figure~\ref{fig:res_train}, a 10-layer residual LSTM showed that its CV cross-entropy does not
degrade with increasing depth. On the contrary, the CV cross-entropy improved. Therefore,
residual LSTMs did not show any training loss observed in~\cite{he2015deep}.
One thing to note is
that the 10-layer residual LSTM also showed 6.7\% training cross-entropy loss.
However, the increased training loss for the residual LSTM network resulted in
better generalization performance like regularization or early-stopping techniques.
It might be due to better representation of input
features from the deep architecture enabled by residual LSTM.

%

\subsection{WER Evaluation with SDM corpus}
\label{sec:exp_fixlayer}
Table~\ref{tab:fixedl} compares WER for plain LSTM, highway LSTM and
residual LSTM with increasing depth. All three networks were trained
by SDM AMI corpus. Both overlapped and non-overlapped
WERs are shown. For each layer, internal memory
cell size is set to be 1024 and output node size is fixed as 512.
A plain LSTM performed worse with increasing layers. Especially, the 10-layer
LSTM degraded up to 13.7\% over the 3-layer LSTM for non-overlapped WER.
A highway LSTM showed better
performance over a plain LSTM but still could not avoid degradation with
increasing depth. The 10-layer highway LSTM presented 6.2\% increase in WER
over the 3-layer network.

On the contrary, a residual LSTM improved with
increasing layers. 5-layer and 10-layer residual LSTMs have 1.2\% and
2.2\% WER reductions over the 3-layer network, respectively.
The 10-layer residual LSTM showed the lowest 41.0\% WER, which
corresponds to 3.3\% and 2.8\% WER reduction over 3-layer plain and
highway LSTMs.

One thing to note is that WERs for 3-layer plain and highway LSTMs are
somewhat worse than results reported in~\cite{zhang2016highway}. The main
reason might be that forced alignment labels used to train LSTM networks are not the
same as the ones used in~\cite{zhang2016highway}. 
1-2\% WER can easily be improved or degraded depending on the quality of
aligned labels. Since the purpose of our evaluation is to measure relative performance
between different LSTM architectures, small absolute difference of WER would not be
any issue. Moreover, reproduce of highway LSTM is based on the open source code provided
by the author in~\cite{zhang2016highway} and therefore, it would be less likely to
have big experimental mismatch in our evaluation.


\subsection{WER Evaluation with SDM and IHM corpora}
\label{sec:exp_increasedC}
Table~\ref{tab:increasedC} compares WER of highway and residual LSTMs trained with combined
IHM and SDM corpora. With increased corpus size, the best performing configuration for a
highway LSTM is changed into 5-layer with 40.7\% WER.  
However, a 10-layer highway LSTM still suffered from training loss from increased depth:
6.6\% increase in WER (non-over).
On the contrary, a 10-layer residual LSTM showed the best WER of 39.3\%,
which corresponds to 3.1\% WER (non-over) reduction
over the 5-layer one, whereas
the prior experiment trained only by SDM corpus presented 1\% improvement.
Increasing training data provides larger gain from a deeper network. Residual LSTM enabled
to train a deeper LSTM network without any training loss.

\begin{table}
\caption{All three LSTM networks have the same size of
layer parameters:1024 memory cells and 512 output nodes. Fixed-size layers
are stacked up when the number of layers increases.
WER(over) is overlapped WER and WER (non-over) is non-overlapped WER.}
\label{tab:fixedl}
\centering
\begin{tabular}{l r r r}
\toprule
 Acoustic Model   & Layer & WER (over) & WER (non-over) \\
 \midrule
	& 3 & 51.1\% & 42.4\% \\
 Plain LSTM   & 5 & 51.4\% & 42.5\% \\
	& 10 & 56.3\% & 48.2\% \\ \midrule
	& 3 & 50.8\% & 42.2\% \\
 Highway LSTM   & 5 & 51.0\% & 42.2\% \\
	& 10 & 53.5\% & 44.8\% \\ \midrule
	& 3 & 50.8\% & 41.9\% \\
 Residual LSTM   & 5 & 50.0\% & 41.4\% \\
	& 10 & 50.0\% & 41.0\% \\
\bottomrule
\end{tabular}
\end{table}

\begin{table}
\caption{Highway and residual LSTMs are trained with combined SDM and IHM corpora.
}
\label{tab:increasedC}
\centering
\begin{tabular}{l r r r}
\toprule
 Acoustic Model   & Layer & WER (over) & WER (non-over) \\
\midrule
	& 3 & 51.3\% & 42.3\% \\
 Highway LSTM   & 5 & 49.5\% & 40.7\% \\
	& 10 & 52.1\% & 43.4\% \\ \midrule
	& 3 & 50.8\% & 41.9\% \\
 Residual LSTM   & 5 & 49.4\% & 40.5\% \\
	& 10 & 48.7\% & 39.3\% \\
\bottomrule
\end{tabular}
\end{table}

\section{Conclusion}
\label{sec:conclusion}
In this paper, we proposed a novel architecture for a deep recurrent neural
network: residual LSTM. Residual LSTM provides a shortcut path between adjacent
layer outputs. Unlike highway network, residual LSTM
does not assign dedicated gate networks for a shortcut connection. Instead,
projection matrix and output gate are reused for a shortcut connection,
which provides roughly 10\% reduction of network parameters compared with
highway LSTMs. Experiments on AMI corpus showed that residual LSTMs
improved significantly with increasing depth, meanwhile 10-layer plain and highway LSTMs
severely suffered from training loss.


\bibliographystyle{IEEEtran}

\bibliography{IEEEabrv,interspeech_main}


\end{document}